%% file: ms.tex
\documentclass[10pt,twocolumn,letterpaper]{article}

\usepackage[pagenumbers]{cvpr} 

\usepackage[pdftex]{graphicx}
\usepackage{amsmath}
\usepackage{amssymb}
\usepackage{booktabs}       

\usepackage{gensymb}
\usepackage[pagebackref,breaklinks,colorlinks]{hyperref}
\usepackage[capitalize]{cleveref}

\crefname{section}{Sec.}{Secs.}
\Crefname{section}{Section}{Sections}
\Crefname{table}{Table}{Tables}
\crefname{table}{Tab.}{Tabs.}

\def\cvprPaperID{11071} 
\def\confName{CVPR}
\def\confYear{2022}

\begin{document}

\title{A study on the distribution of social biases in self-supervised learning visual models}

\author{
Kirill Sirotkin,  Pablo Carballeira, Marcos Escudero-Viñolo \thanks{This work was supported by the Consejería de Educación e Investigación of the Comunidad de Madrid under Project SI1/PJI/2019-00414} \\
Universidad Autónoma de Madrid, 28049, Madrid, Spain\\
{\tt\small\{kirill.sirotkin, pablo.carballeira, marcos.escudero \}@uam.es} 
}

\maketitle
\input{Abstract}
\input{Introduction}
\input{Related_work}
\input{Methodology}

\input{Results}

\input{Discussion}
\input{Conclusion}

{\small
\bibliographystyle{ieee_fullname}
\bibliography{Bibli}
}

\end{document}

%% file: Abstract.tex
\begin{abstract}

Deep neural networks are efficient at learning the data distribution if it is sufficiently sampled. However, they can be strongly biased by non-relevant factors implicitly incorporated in the training data. These include operational biases, such as ineffective or uneven data sampling, but also ethical concerns, as the social biases are implicitly present—even inadvertently, in the training data or explicitly defined in unfair training schedules. In tasks having impact on human processes, the learning of social biases may produce discriminatory, unethical and untrustworthy consequences. It is often assumed that social biases stem from supervised learning on labelled data, and  thus, Self-Supervised Learning (SSL) wrongly appears as an efficient and bias-free solution, as it does not require labelled data. However, it was recently proven that a popular SSL method also incorporates biases. In this paper, we study the biases of a varied set of SSL visual models, trained using ImageNet data, using a method and dataset designed by psychological experts to measure social biases. We show that there is a correlation between the type of the SSL model and the number of biases that it incorporates. Furthermore, the results also suggest that this number does not strictly depend on the model's accuracy and changes throughout the network. Finally, we conclude that a careful SSL model selection process can reduce the number of social biases in the deployed model, whilst keeping high performance.

\end{abstract}

%% file: Introduction.tex
\section{Introduction} \label{section_intro}

Supervised Deep Learning models currently constitute the state-of-the-art in the fields of computer vision and natural language processing. However, the recent developments \cite{bib_byol, bib_moco_v3} in the field of Self-Supervised Learning (SSL) \textemdash a type of unsupervised learning, are slowly closing the performance gap gained via human guidance usually provided in the shape of target labels. SSL methods aim at solving a pre-formulated pretext task\textemdash defined by automatically generated labels, whose solution is expected to require high-level understanding of the data in order to learn descriptive feature embeddings with strong transferability potential.

Human social biases are a well-studied and, in some cases, numerically quantifiable phenomenon \cite{bib_iat} that causes unjustified prejudices against social groups based, among others, on aspects such as age, gender and race. Whereas one can not assign prejudices or preferences to deep learning approaches as these are highly subjective characteristics attributed solely to humans, deep learning methods can wrongly correlate certain concepts if the labeled training data distribution is biased itself \cite{bib_bias_word5}. In practice, this leads to the replication of social biases. Several cases have been studied and reported, including: an  incorrect gender prediction based on the contextual cues (i.e. location - kitchen, office), rather than on visual evidence associated with the described person \cite{bib_snowboard}, 
a fewer number of automatic high-paying job recommendations for female candidates than for male ones \cite{bib_job_ads} and a promotion of biased suggestions in the dating/political decision-making context \cite{bib_dating_bias}. Anticipating these situations, institutional initiatives are being developed internationally to extinguish social biases from the training data, as declared in the Ethics Guidelines for a Trustworthy AI issued by the European Commission, and regulate the use of machine learning methods with potential human implications, as stated in numerous US bills \cite{bib_sb_6280, bib_sb_S5140B, bib_sf_ban} and the legislative documents of other countries \cite{bib_jamaica, bib_india, bib_kenya, bib_uk}.

Previously, it was demonstrated that supervised learning models are prone to implicitly learn biases from the datasets containing them \cite{bib_snowboard, bib_biased_data, bib_equality}, as these human biases are encapsulated in the target labels. For instance, it has been shown that the earlier versions of ImageNet \cite{bib_imagenet} exposed an imbalanced distribution regarding skin colors, ages and genders, leading to the under-representation of certain groups \cite{bib_bad_imagenet}. Furthermore, datasets collecting raw comments scraped from the web \cite{bib_comments_dataset, bib_toxic_comments} contain explicit biases against certain social groups \cite{bib_disability}.

SSL approaches, being unsupervised, are expected to be unaffected by biases-bearing labels. However, as they require large amounts of training data that often prevents its curation, it is not unlikely that the data itself contains some social human biases. In fact, results for a recent study \cite{bib_biases_ieat}, suggest that two of the state-of-the-art unsupervised learning models also contain association biases learned from the data, only in this case it can not be explained by the choice of class labels, as the unsupervised models do not leverage this information in the training process. This result indicates that at least one top performing SSL model \cite{bib_simclr_v2} might implicitly learn social biases while training to solve the targeted pretext task. Hence, data should be handled carefully, as the neural network's capacity to avoid inadvertent perpetuation of undesirable social biases is an important quality to consider, alongside classification accuracy, in the design of deep learning models. 




This papers addresses this phenomenon and attempts to answer the following questions: what is the origin of the biases in the SSL setting? What affects the model's proneness to learn an implicit social bias? What is the relationship between the model's accuracy and the biases it learns? Whereas a preliminary work  addresses the first question and hypothesizes on the origins of implicit social biases in a couple of unsupervised models \cite{bib_biases_ieat}, to our knowledge, this is the first attempt in studying a wider and more varied set of SSL models. 

In particular, the contributions of this paper are:
\begin{itemize}

\item We study the association biases acquired by 11 SSL models that share the same ResNet-50~\cite{bib_resnet} architecture, and vary in terms of pretext task and, thus, their accuracy after transfer learning. The results of this study suggest that the nature of the pretext task influences the number and nature of incorporated biases, and that contrastive models are more prone to acquire  biased associations that are implicit in the data.
    
\item We also perform an analysis of biases acquired in the embeddings at different layers of the models, showing that the number and strength of the biases vary at different model depths.  The results of the per-layer analysis suggest that a careful consideration of bias in transfer learning applications can improve the trade-off between bias and accuracy, as the accuracy achieved using embeddings from highly-biased layers is not far from the accuracy achieved by a less-biased embeddings layer.
    
    
    
\end{itemize}


%% file: Related_work.tex
\section{Related work} \label{section_relat}

\subsection{Measuring biases of computer vision models} \label{section_rel_biases}


Many of the existing methods for measuring the biased associations are based on the Implicit Association Test (IAT) \cite{bib_iat} that measures the differential relationships between a target concept and an attribute. The IAT measures the difference in reaction time of a respondent when correlating concepts and attributes for which biased associations are prone to exist and for which they are not. For instance, a biased test subject strongly associating a concept \textit{flower} with an attribute \textit{pleasant} takes less time to correlate verbal or visual stimuli representing them rather than correlating stimuli representing a concept \textit{insect} with an attribute \textit{pleasant}.



Until recently, association bias tests were mostly used for Natural Language Processing (NLP) \cite{bib_bias_word_embeddings, bib_bias_words2, bib_bias_words3, bib_bias_words4, bib_bias_word5}, but a recent work has extended the Word Embedding Association Test (WEAT) \cite{bib_biases} to the image domain, thereby, making it possible to quantify association biases in computer vision models. This approach, named Image Embedding Association Test (iEAT), measures the differential association of the target concepts X and Y with the attributes A and B based on the image embeddings obtained by feeding images representing these concepts and attributes to a trained deep learning model. For instance, let the chosen target concepts be \textit{insect} (X) and \textit{flower} (Y) and the attributes be \textit{unpleasant} (A) and \textit{pleasant} (B). Then, the association test will measure the strength of correlation between \textit{insect} and \textit{unpleasant}, and \textit{flower} and \textit{pleasant} based on the cosine distances between the embeddings of X, Y, A and B. A more detailed explanation is given in Section~\ref{section_methodology}.

\subsection{Image embeddings via Self-supervised learning models} \label{section_SSL_types}

In this paper, we refer to an image embedding as the features extracted at a given layer of a deep learning model when a particular image is fed to it. These embeddings are accepted as a representative description of the image \textemdash subjected to the training target. Usually, one can expect that, at a given layer and for a given architecture, the higher the performance of the learned model is, the more representative the embeddings will be. A common way to obtain image embeddings is by using a network trained in the supervised mode \cite{bib_resnet, bib_inception}. Alternatively, SSL models can be used if images are to be represented in label scarce scenarios\textemdash as medical data requiring expert annotations or data acquired using devices capturing at non-visual modalities. SSL methods, instead of being trained for a label-driven task can be trained by using objectives such as a simple geometric task \cite{bib_rotation, bib_relative_location}, pseudo labels generated through automatic clustering \cite{bib_dc, bib_odc}, or promoting proximity of "similar" data points in the feature space \cite{bib_byol, bib_moco_v3, bib_simclr_v2, bib_moco_v1, bib_moco_v2, bib_simclr}. These objectives are commonly known as pretext tasks and can be used to arrange SSL models into the following three groups.

\begin{figure*}[t]
  \centering
 \begin{subfigure}[b]{0.07\textheight}
      \includegraphics[width=\textwidth]{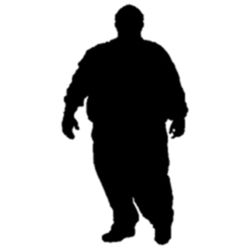}
 \end{subfigure}
  \hfill
  \begin{subfigure}[b]{0.07\textheight}
      \includegraphics[width=\textwidth]{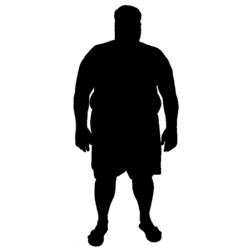}
 \end{subfigure}
  \hfill
  \begin{subfigure}[b]{0.07\textheight}
      \includegraphics[width=\textwidth]{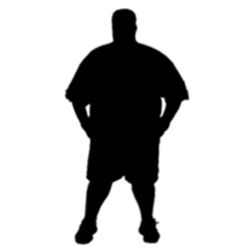}
 \end{subfigure}
  \hfill
  \begin{subfigure}[b]{0.07\textheight}
      \includegraphics[width=\textwidth]{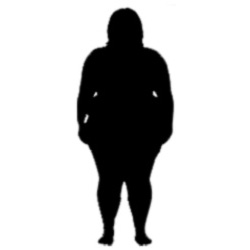}
  \end{subfigure}
  \hfill
  \begin{subfigure}[b]{0.07\textheight}
      \includegraphics[width=\textwidth]{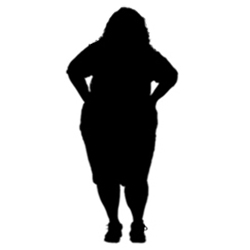}
  \end{subfigure}
  \hfill
  \begin{subfigure}[b]{0.07\textheight}
      \includegraphics[width=\textwidth]{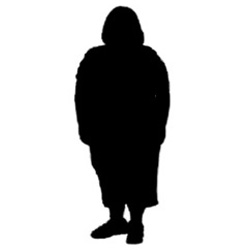}
  \end{subfigure}
  \vfill
  \begin{subfigure}[b]{0.07\textheight}
      \includegraphics[width=\textwidth]{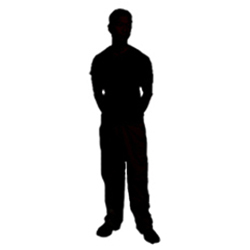}
  \end{subfigure}
  \hfill
  \begin{subfigure}[b]{0.07\textheight}
      \includegraphics[width=\textwidth]{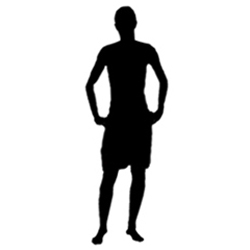}
  \end{subfigure}
  \hfill
  \begin{subfigure}[b]{0.07\textheight}
      \includegraphics[width=\textwidth]{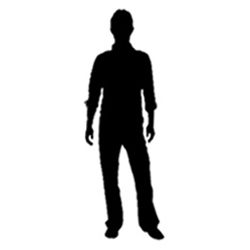}
  \end{subfigure}
  \hfill
  \begin{subfigure}[b]{0.07\textheight}
      \includegraphics[width=\textwidth]{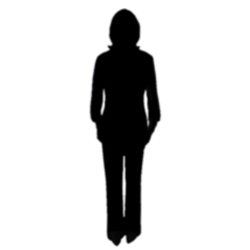}
  \end{subfigure}
  \hfill
  \begin{subfigure}[b]{0.07\textheight}
      \includegraphics[width=\textwidth]{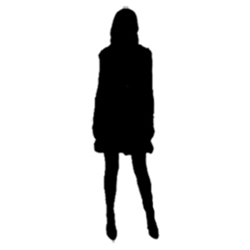}
  \end{subfigure}
  \hfill
  \begin{subfigure}[b]{0.07\textheight}
      \includegraphics[width=\textwidth]{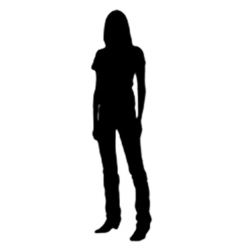}
  \end{subfigure}
  \caption{Images representing the concepts of overweight people (at the top) and thin people (on the bottom). The images are used to identify a weight-valence bias in the iEAT and original IATs \cite{bib_iat}.}
  \label{fig_fat_thin}
\end{figure*}

\begin{figure}[h]
  \centering
 \begin{subfigure}[l]{0.07\textheight}
      \includegraphics[width=\textwidth]{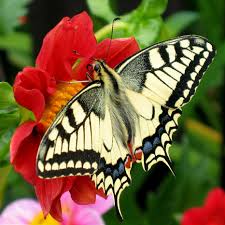}
 \end{subfigure}
  \hfill
  \begin{subfigure}[l]{0.07\textheight}
      \includegraphics[width=\textwidth]{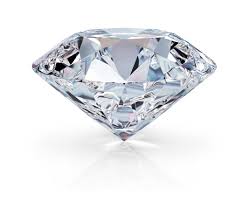}
 \end{subfigure}
  \hfill
  \begin{subfigure}[l]{0.07\textheight}
      \includegraphics[width=\textwidth]{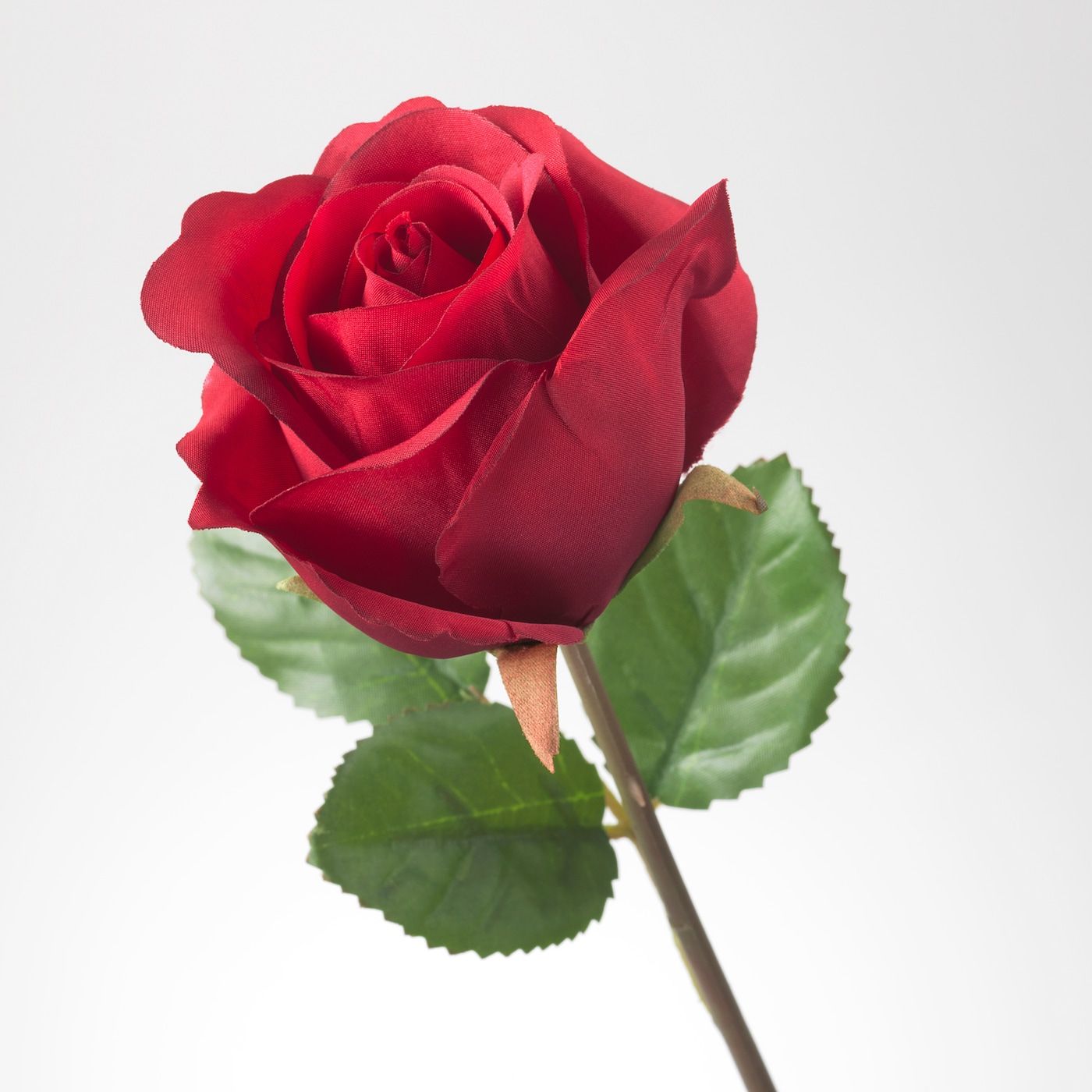}
 \end{subfigure}
\hfill
  \begin{subfigure}[l]{0.07\textheight}
      \includegraphics[width=\textwidth]{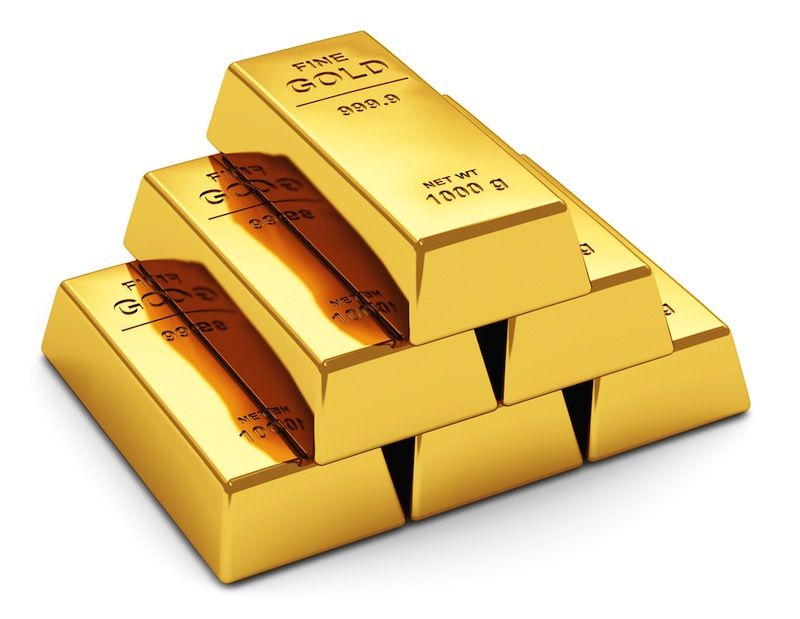}
  \end{subfigure}
  \hfill
  \begin{subfigure}[l]{0.07\textheight}
      \includegraphics[width=\textwidth]{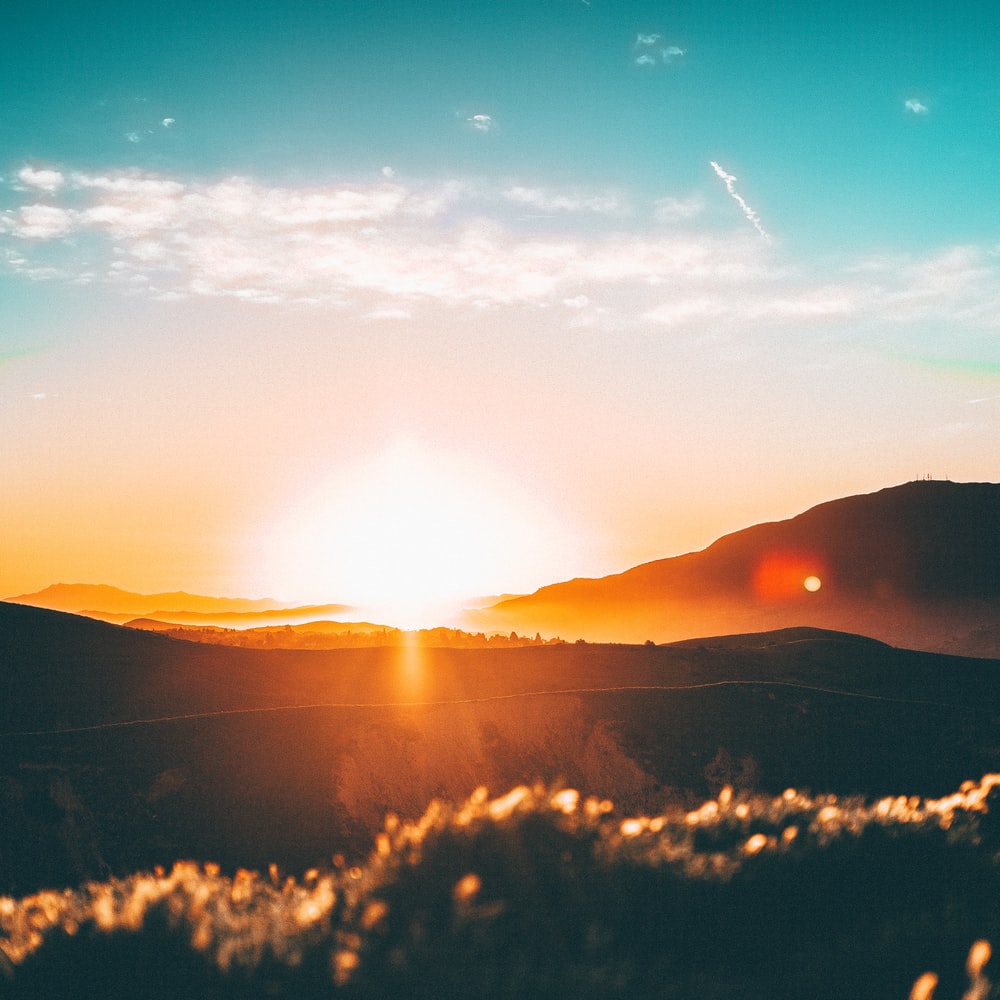}
  \end{subfigure}
  \vfill
  \begin{subfigure}[l]{0.07\textheight}
    \includegraphics[width=\textwidth]{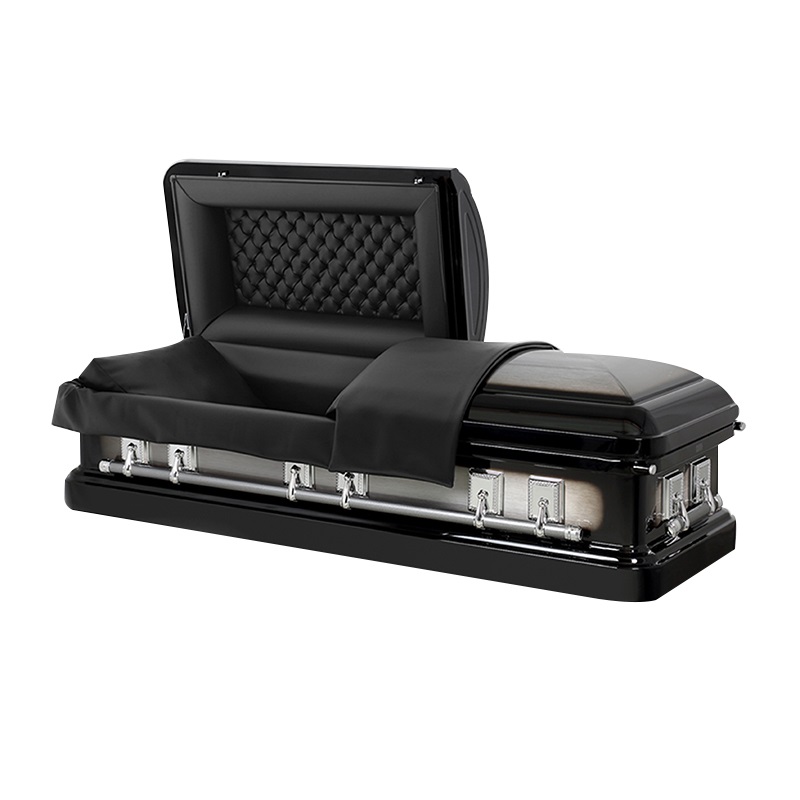}
  \end{subfigure}
  \hfill
  \begin{subfigure}[l]{0.07\textheight}
      \includegraphics[width=\textwidth]{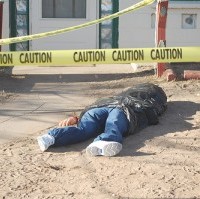}
  \end{subfigure}
  \hfill
  \begin{subfigure}[l]{0.07\textheight}
      \includegraphics[width=\textwidth]{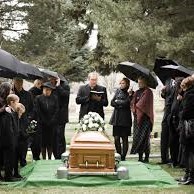}
  \end{subfigure}
  \hfill
  \begin{subfigure}[l]{0.07\textheight}
      \includegraphics[width=\textwidth]{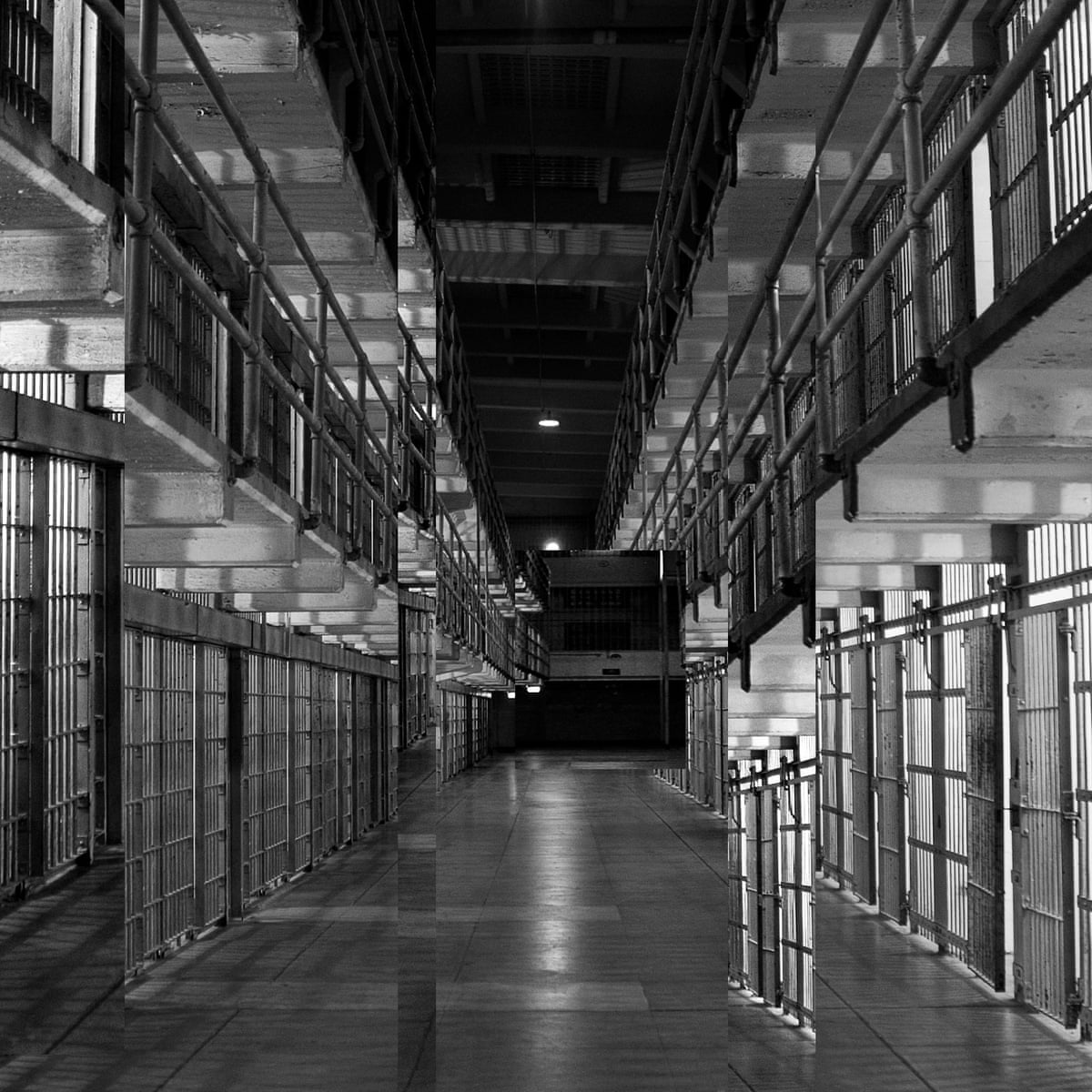}
  \end{subfigure}
  \hfill
  \begin{subfigure}[l]{0.07\textheight}
      \includegraphics[width=\textwidth]{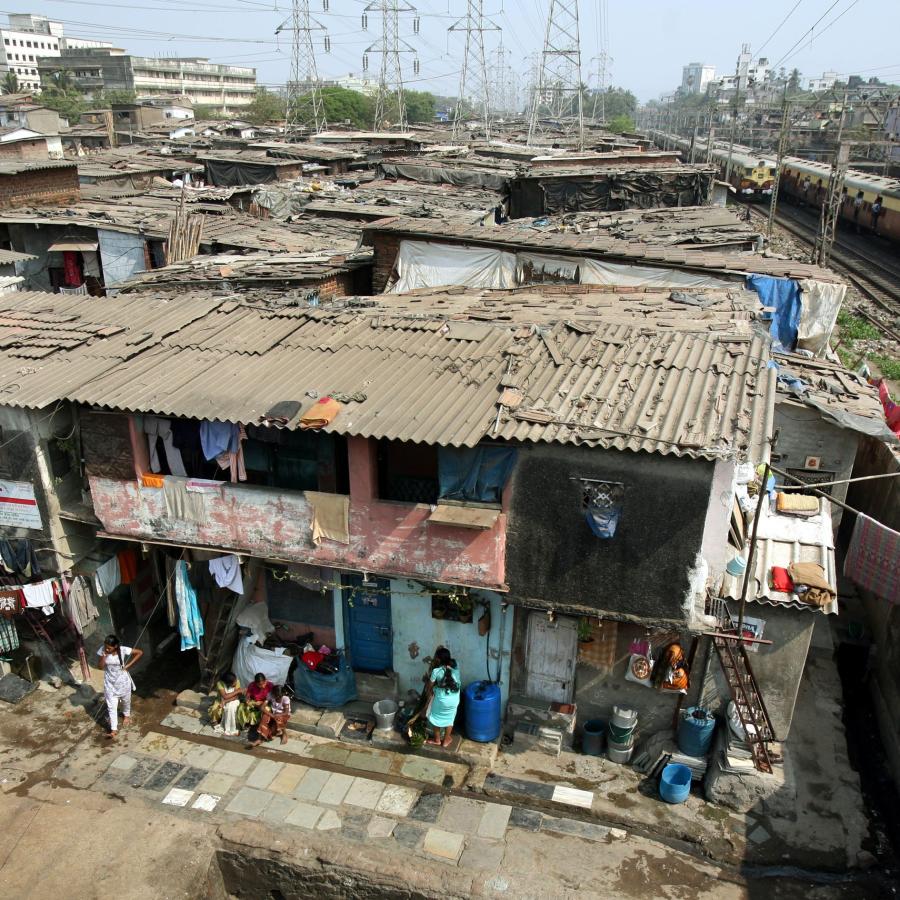}
  \end{subfigure}
  \caption{Images representing the concepts of valence: pleasant (at the top) and unpleasant (on the bottom). The images correspond to the verbal stimuli commonly used to describe the valence concepts \cite{bib_iat}.}
  \label{fig_valence}
\end{figure}

\paragraph{Geometric models} \label{sub_geometric}   
One of the most straightforward approaches to defining a pretext task is applying a geometric transformation to an input image and training a network to solve it. The three geometric pretext tasks considered in this paper are rotation prediction \cite{bib_rotation}, relative patch location prediction \cite{bib_relative_location} and jigsaw puzzles \cite{bib_jigsaw}. The rotation prediction pretext task randomly applies one out of 4 rotations: 0\degree, 90\degree, 180\degree, 270\degree, to each training image sample and trains the network to predict which rotation was applied to a given image. On the other hand, a model trained to predict patch locations is based on randomly sampling two close regions from an input image and training the network to predict their relative spatial location. Finally, when a jigsaw puzzle strategy is followed, the image is divided into tiles, that are then randomly shuffled. Then, the network is trained to predict their original arrangement.

\paragraph{Clustering-based representation learning} \label{sub_clustering}
A more sophisticated approach to deep unsupervised learning is based on the classical clustering methods that are used to group unlabeled data into clusters according to some homogeneity criteria. An obvious way to incorporate clustering into the pretext task formulation is to perform clustering after each model update step. The generated labels are then used as pseudo-labels to evaluate the model in a supervised manner. These labels would, in turn, change the embeddings at the next step as the newly generated labels may differ from the labels at the previous step. This is the strategy followed by Deep Clustering (DC) \cite{bib_dc}, that suffers from instability during the training process due to the random permutation of labels at each step. To tackle the issue of labels permutation and instability, Cluster Fit \cite{bib_clusterfit} relies on using a teacher network to define the pseudo-labels. Differently, in Online Deep Clustering (ODC) \cite{bib_odc} the labels are updated using mini-batches and this process is integrated into the model update. This way, the embeddings and labels evolve together and the instability inherent in DC is eliminated.

\paragraph{Contrastive models} \label{sub_contrastive}
Top performing SSL models are driven by pretext tasks using contrastive losses \cite{bib_contrastive_loss}. Although exact implementations vary from model to model, the main idea remains the same: to learn representations that map the \textit{positives} close together and push apart the \textit{negatives}. The \textit{positive} samples might be chosen based on modifications of patches in the same image or applying different augmentations obtained from the same image. 

Non-Parametric Instance Discrimination (NPID) \cite{bib_npid} treats each input image (instance) as belonging to a unique class and trains the classifier to separate between each instance via the noise-contrastive estimation \cite{bib_noise_contrastive}. The motivation for it comes from the observation that supervised learning approaches return similar embeddings for related images. Specifically, it is often the case that the second top scoring predicted class at the end of the model is semantically close to the first one following a human interpretation. Therefore, the network is expected to learn the semantic similarity between classes without explicitly having it as the objective.

Momentum Contrast (MoCo) \cite{bib_moco_v1} leverages a dynamic dictionary where \textit{query} and associated \textit{keys} represent image encodings obtained with an encoder network. If a  \textit{query} and a  \textit{key} come from the same image, they are considered to be a  \textit{positive} pair, otherwise a  \textit{negative} one. The  \textit{queries} and the  \textit{keys} are encoded by separate networks and the  \textit{key} encoder is updated as a moving average of the  \textit{query} encoder, enabling a large and consistent dictionary for learning visual representations.

Simple Framework for Contrastive Learning of Visual Representations (SimCLR) \cite{bib_simclr}, building on the principles of contrast learning, introduces a series of design changes that allow it to outperform MoCo \cite{bib_moco_v1} not requiring a memory bank. Among these changes are a more careful choice of data augmentation strategies, addition of a non-linearity between the embeddings and the contrastive loss, and increased batch sizes and the number of training steps. Further improving on the results of SimCLR \cite{bib_simclr}, the second version of Momentum Contrast model (MoCo v2) \cite{bib_moco_v2} acknowledges its efficient design choices and takes advantage of an MLP projection head and more data augmentations.

Bootstrap Your Own Latent (BYOL) \cite{bib_byol} reaches a new state-of-the-art on ImageNet linear classification while avoiding one of the greatest challenges that other contrastive models face: a need for negative pairs. BYOL circumvents this problem by generating the target representations with a randomly initialized model and then using them for its online training, by iteratively updating the target network, the online network is expected to learn better and better representations.

Finally, SwAV \cite{bib_swav} describes a hybrid clustering-contrastive method that avoids the computation of pairwise distances between positive and negative samples by clustering the data in consistency-enforced clusters of the different image augmentations. Thereby, defining positive samples according to cluster memberships and reducing the distance storage requirements of the other contrastive methods.

%% file: Methodology.tex
\section{Methodology} \label{section_methodology}

\setlength{\tabcolsep}{1pt}
\begin{table*}[]
\centering
\caption{Hyperparameters used to train the SSL models and the fully supervised ResNet-50.}
\label{table_hyperparams}
\begin{tabular}{@{}ccccccccccccc@{}}
\toprule
           & Jigsaw & RL  & ClusterFit & Rotation & NPID & ODC  & MoCo v1 & SimCLR & MoCo v2 & BYOL & SwAV & Sup. \\ \midrule
Batch size & 256    & 512 & 256        & 512      & 256  & 512  & 256     & 4096   & 256     & 4096 & 4096 & 256  \\
Epochs     & 105    & 70  & 105        & 70       & 200  & 440  & 200     & 200    & 200     & 200  & 200  & 90   \\
Base lr    & 0.1    & 0.2 & 0.1        & 0.2      & 0.03 & 0.06 & 0.03    & 0.3    & 0.03    & 0.3  & 0.3  & 0.1  \\
\begin{tabular}[c]{@{}c@{}}ImageNet accuracy\\ (best layer)\end{tabular} & 48.57 & 49.31 & 53.63 & 54.99 & 56.61 & 57.70 & 61.02 & 66.61 & 67.69 & 71.61 & 73.85 & 74.12 \\ \bottomrule
\end{tabular}
\end{table*}

Given the set of SSL models described in Section \ref{section_SSL_types}, we apply the iEAT framework \cite{bib_biases_ieat} introduced in Section \ref{section_rel_biases} to each model and investigate the presence of association biases. iEAT takes a set of input embeddings  $\lbrace x, y \rbrace$ of images representing the target concepts $(X, Y)$ and a set of input embeddings $\lbrace a, b \rbrace$ representing the measured attributes $(A, B)$. For instance, the target concepts \textit{overweight people} and \textit{thin people} are represented by the example images on Figure~\ref{fig_fat_thin}, while the attributes \textit{pleasant} and \textit{unpleasant} can be visualized by the images representing the \textit{valence} concept (see examples in Figure~\ref{fig_valence}). The null-hypothesis tested by iEAT states that $X$=\textit{overweight people}  embeddings are as similar as  $Y$=\textit{thin people} embeddings to $(A, B)$=$($\textit{pleasant}, \textit{unpleasant}$)$ embeddings, or that the dissimilarity is equally small. The rejection of the null-hypothesis would mean that one target concept is more correlated with one attribute than the other target concept, thus, detecting an association bias. iEAT tests the null-hypothesis by a permutation test and a metric quantifying the differential association $s(X,Y,A,B)$, defined as follows:

\setlength{\abovedisplayskip}{0pt}
\setlength{\belowdisplayskip}{0pt}
\setlength{\abovedisplayshortskip}{0pt}
\setlength{\belowdisplayshortskip}{0pt}
\begin{equation}
s(X,Y,A,B) = \sum_{x\in X} s(x,A,B) - \sum_{y\in Y} (y,A,B),\label{eq_ieat}
\end{equation}

where:

\begin{equation}
    s(t,A,B) = \underset{a\in A}{mean} \cos(t,a) - \underset{b\in B}{mean} \cos(t,b) \text{ for } t=\lbrace x,y \rbrace. \label{eq_cos}
\end{equation}



The permutation test randomly shuffles the labels of the set of embeddings representing target concepts $(X,Y)$, creating 10000 randomly permuted sets (or the maximum number of permutations allowed by the set size). Then, the differential association (Eq. \ref{eq_ieat}) of each one of these permuted sets is measured. The $p$-value collects the percentage of permuted sets resulting in a larger or equal differential association than the original set. The null-hypothesis can be rejected (and thus a bias detected) with high probability if the $p$-value is below a certain threshold. The strength of the bias can be measured as the effect size ($d$-value) \textemdash measures the separation between the two distributions of the distances of the two sets of target concept samples to the attribute samples~\cite{bib_bias_words2}:

\begin{equation}
d = \frac{ \underset{x\in X}{mean} \text{ } s(x,A,B) - \underset{y\in Y}{mean} \text{ } s(y,A,B)}{ \underset{t\in X \cup Y}{std} s(t,A,B)}.\label{eq_effect_size}
\end{equation}  




The full bias-detection pipeline for a given network model, therefore, consists of the extraction of deep feature embeddings of 4 image sets representing 2 target concepts and 2 attributes (i.e. Office-Home vs. Male-Female) with the same model, and running the permutation test described above on these embedding sets. 



We evaluate the SSL models on the data provided by the authors of the iEAT framework\footnote{distributed under CC BY-NC-SA 4.0 license} \cite{bib_biases_ieat}. This data encompasses visual stimuli for sets of target concepts such as race, gender and age. The dataset contains 3 to 55 psychologists-selected images per concept taken from well-established IAT tests \cite{bib_iat}, CIFAR-100 dataset \cite{bib_cifar_100} or the web. Following the approach outlined above, we collect the $p$- and $d$-values representing the likelihood and strength of each of the 39 association biases proposed in the iEAT framework (see supplementary material for a complete list).


All models used in this paper were trained on ImageNet2012 \cite{bib_imagenet} and share the same backbone architecture: ResNet-50\cite{bib_resnet}. The weights for the pretrained networks were taken from the \href{https://github.com/open-mmlab/OpenSelfSup/blob/master/docs/MODEL_ZOO.md}{OpenSelfSupervised Framework}\footnote{distributed under Apache License 2.0} and \href{https://github.com/facebookresearch/vissl/blob/main/MODEL_ZOO.md}{VISSL}\footnote{distributed under MIT License} with training hyper-parameters listed in Table~\ref{table_hyperparams}.  We evaluate embeddings obtained from the first max pooling layer (layer 1 hereinafter), as well as the embeddings obtained after each ResNet block (layers 2-5) and the final Global Average Pooling (GAP). To achieve a more comprehensive overview of the presence of biases in the SSL models, we do not limit our choice to the state-of-the-art architectures and select networks that conceptually represent different approaches for SSL: Rotation prediction (Rotation) \cite{bib_rotation}, Relative patch location prediction (Relative Location, RL) \cite{bib_relative_location}, Jigsaw puzzles \cite{bib_jigsaw}, SwAV \cite{bib_swav}, ClusterFit \cite{bib_clusterfit},
ODC \cite{bib_odc}, NPID \cite{bib_npid}, MoCo\_v1 \cite{bib_moco_v1}, MoCo\_v2 \cite{bib_moco_v2}, SimCLR \cite{bib_simclr} and BYOL \cite{bib_byol}, as well as a randomly initialized ResNet-50 (random) and a fully supervised ResNet-50 (supervised) \cite{bib_resnet}. 






%% file: Results.tex
\section{Experimental results} \label{section_res}

\begin{figure}[!tbp]
  \centering
  \begin{minipage}[t]{0.49\textwidth}
    \includegraphics[width=\textwidth]{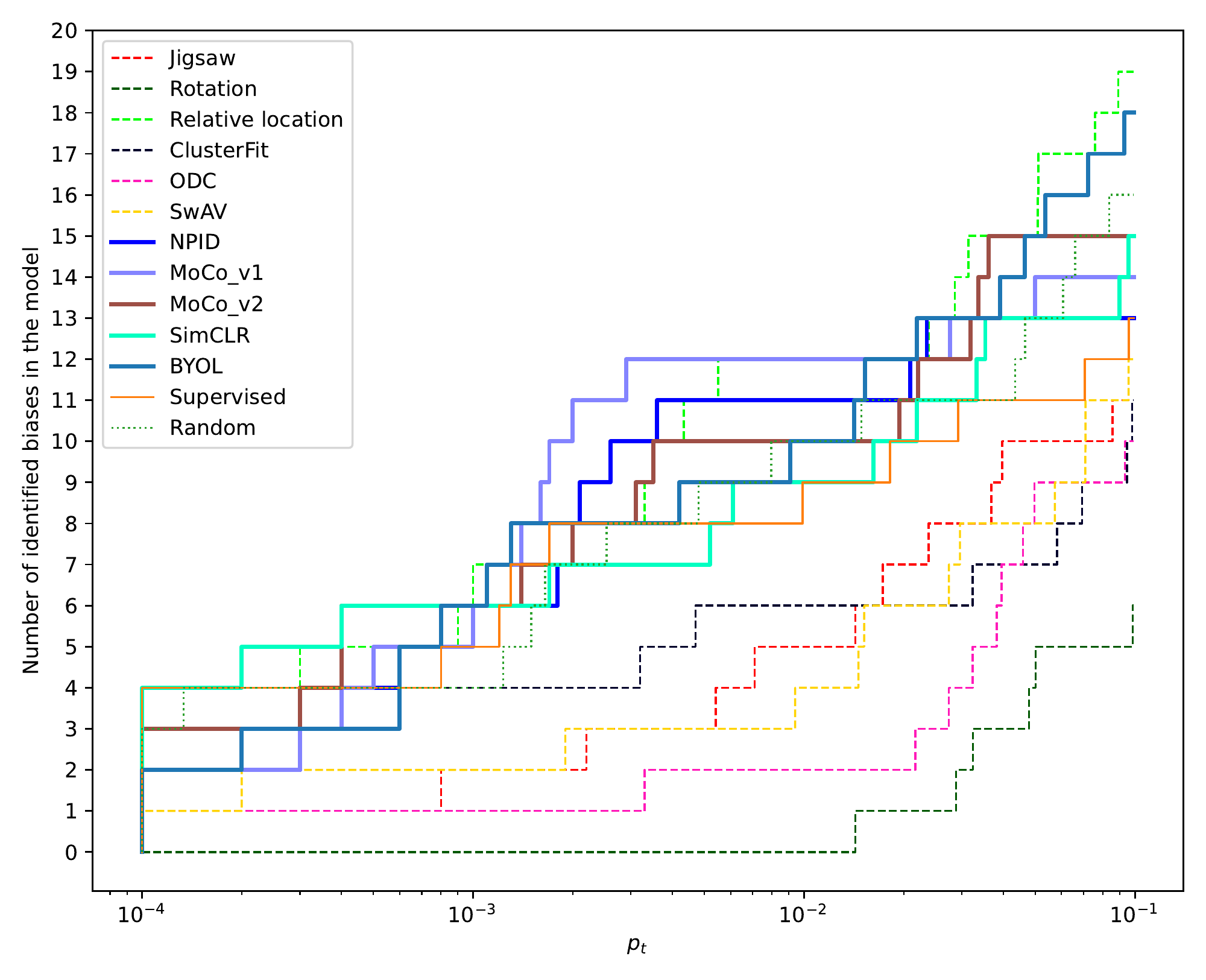}
    \caption{Number of biases detected in the embeddings of the Global Average Pooling layer for different values of $p_t$. Biases detected for lower values of $p_t$ are statistically more significant. Contrastive models are plotted with thick solid lines, geometric models and clustering-based models with dashed lines.}
    \label{fig_cumulative_avg_pool}
  \end{minipage}
  \hfill
  \begin{minipage}[t]{0.49\textwidth}
    \includegraphics[width=\textwidth]{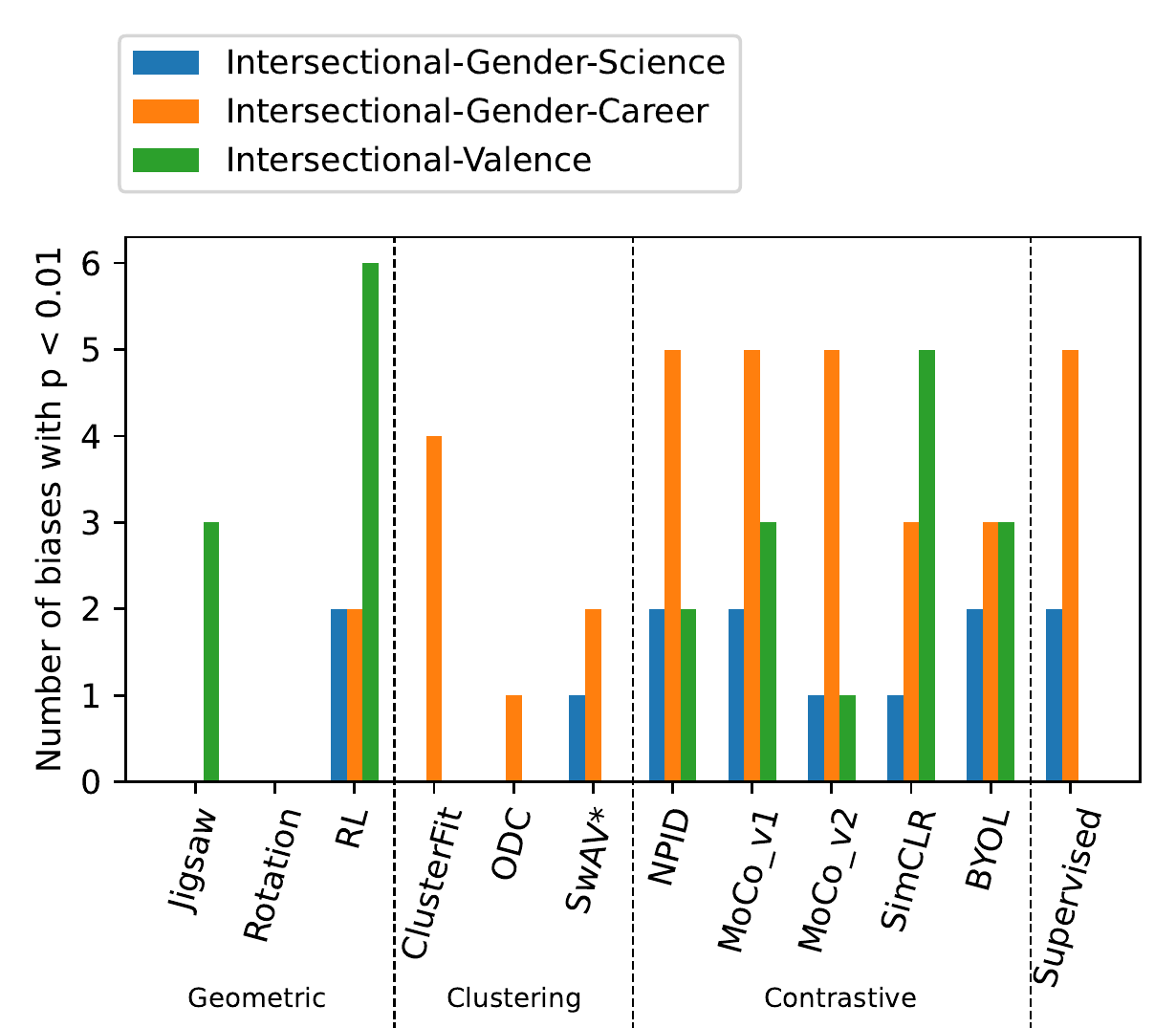}
    \caption{Numbers of intersectional biases detected in the embeddings of the Global Average Pooling layer with $p_t < 0.01$. Note that a hybrid clustering-contrastive model SwAV\cite{bib_swav} is labeled as a clustering method for better readability of the figure.}
  \label{fig_intersectional_bars}
  \end{minipage}
\end{figure}


This section summarizes the results of the bias detection on the deepest ResNet-50 embeddings, as well as on embeddings of its intermediate and shallow layers. 

As stated in Section~\ref{section_methodology}, the $p$-values are obtained with the permutation test. Due to the inherent randomness of this test, we repeat each experiment three times to confirm the consistency of the results and average the obtained $p$- and $d$-values (statistical error results are provided in the supplementary material). Moreover, we evaluate three instances of the random ResNet-50 model, in order to account for the random weight initialization. Thus, the $p$- and $d$-values reported for the random model are averaged for three instances and three permutation tests.

\subsection{Bias-analysis for the GAP embeddings}



The first analysis of the presence of social biases on different SSL models is performed using the embeddings at the deepest layer of the CNN architecture, i.e. after the GAP layer of ResNet-50, (GAP embeddings hereafter). GAP embedings generally convey a high performance in transfer learning scenarios (although not necessarily the highest). Thus, given the reduced dimensionality of the GAP embeddings with respect to those extracted at previous layers (which leads to faster training of a classifier on top of them) they are a common choice for transfer learning applications.

A bias is considered to be present in the model if its statistical significance, measured with the $p$-value yielded by the permutation test, is below a certain threshold $p_t$. Given the absence of a universally correct $p_t$ value, a possible approach to categorize bias detection is to split them into groups of significance, as done in previous works~\cite{bib_biases_ieat}. Here, we take a similar approach, and explore the biases acquired by a model for $p_t$ values in the $[10^{-4},10^{-1}]$ range \textemdash which sits in the range of high statistical significance.
Figure \ref{fig_cumulative_avg_pool} shows the number of biases found at the GAP embeddings of the thirteen considered models with respect to the $p$-value threshold. From Figure \ref{fig_cumulative_avg_pool} one can observe a clear separation in the number of detected biases between two groups of models: contrastive SSL models yield more biases than geometric/clustering-based models, with the exception of the relative location model. For example, at $p_t = 10^{-2}$ no bias is detected for the rotation model and only 2 biases are detected for ODC, while the number of biases for contrastive models ranges from 8 to 12. This holds for any value of the threshold, showing the reliability of this conclusion. This is also shown on the large differences in the number of acquired biases among the three groups of models depicted in Figure~\ref{fig_intersectional_bars}. Especially for the \textit{intersectional} biases (the most common ones) detected at the GAP embeddings. Results for all embeddings and biases are in the supplementary material.

\subsubsection{Contrastive vs Non-Contrastive models}
We quantitatively assess the difference in the number of biases acquired by contrastive and non-contrastive models, by a statistical analysis of the the biases present in two equal-sized  sets of models: i) contrastive (NPID, SimCLR, MoCo\_v1, MoCo\_v2, BYOL) and ii) non-contrastive (Jigsaw, Rotation, ODC, Relative location, ClusterFit):

\begin{enumerate}
    \item Initially, $p_t$ is set to 0, and is gradually increased to 0.1 (i.e., $p_t$ is moved in the direction of the x-axis of Figure~\ref{fig_cumulative_avg_pool}). 
    
    \item For each $p_t$ value, we: 
    \begin{enumerate}
        \item Compute $\delta_{orig}$: difference between the number of biases for the contrastive and non-contrastive sets.

        \item Permute the labels between the contrastive and non-contrastive sets, and, for each permutation $k$, compute $\delta_{k}$: the difference between the number of biases for the two sets generated in the permutation.

        \item For each permutation $k$, check if ($\delta_{orig} < \delta_{k}$), and use this to estimate the probability of randomly permuted sets having more biases than the original sets.
        
    \end{enumerate}

    \item Finally, we average the probabilities computed for each step of $p_t$.
\end{enumerate}

This test yields a $p$-value of 0.045 that validates the premise by the rejection of the null-hypothesis: "contrastive models \textit{are not more biased} than other models".


Finally, if we compare the bias data in Figure \ref{fig_cumulative_avg_pool} with the accuracy data in Table~\ref{table_hyperparams}, there is no  direct link between the model's classification accuracy and the number of biases it incorporates. Indeed, BYOL and MoCo acquired more biases than the more accurate supervised model. Moreover, the least accurate model (relative location) is one of the models that incorporates the highest number of biases.


\subsection{Bias detections in the random model} \label{bias random}

Performing the bias analysis on the embeddings of the baseline random  ResNet-50 model, we discovered a high number of biases. While it is implausible that a randomly initialized model can consistently contain certain biases, the bias detections themselves are possible in conditions that relate to the specific test data. We hypothesize that the bias detections in the random models come from the correlations in the test data caused by strong similarities between some low-level features. To test this hypothesis we randomly permute the pixels in the images representing two target concepts (i.e. Weapon, Tool), while leaving the images representing the two attributes (i.e. Black, White) unchanged, and repeat the bias test. This allows to remove the high-level visual concepts and most of the low-level features (such as textures) from the images, preserving only the distribution of pixel values. We perform this test for 13 social biases that are detected in the random model and observe that after the permutation of pixels, 11 of them remain present (see supplementary material for complete results) and, in some cases, even have lower $p$-values (Lincoln-Trump vs. Pleasant-Unpleasant).

\begin{figure*}[t]
  \centering
 \begin{subfigure}[b]{0.49\textwidth}
       \includegraphics[width=\textwidth]{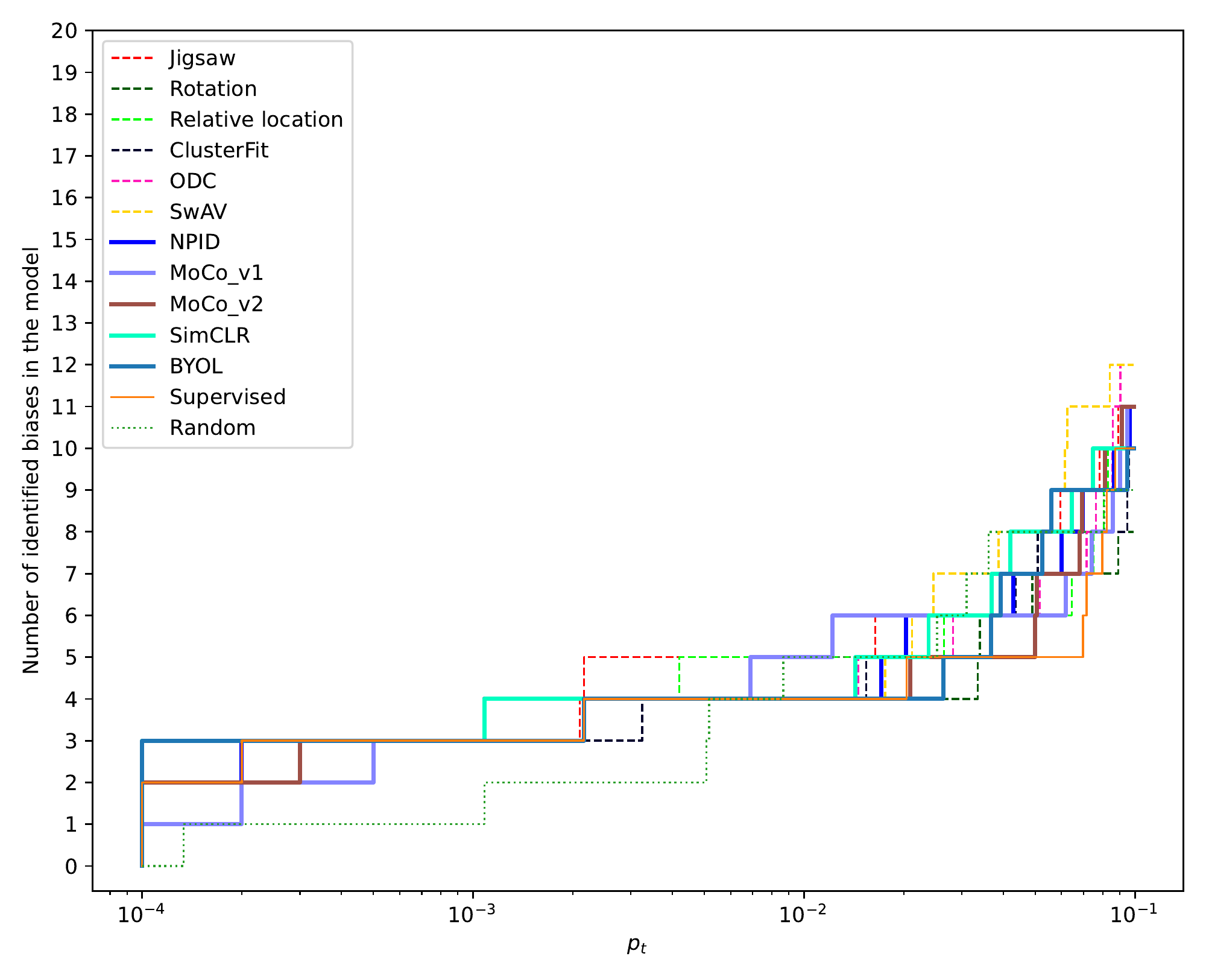}
       \caption{2\textsuperscript{nd} ResNet block}
       \label{fig_2a}
 \end{subfigure}
  \hfill
  \begin{subfigure}[b]{0.49\textwidth}
       \includegraphics[width=\textwidth]{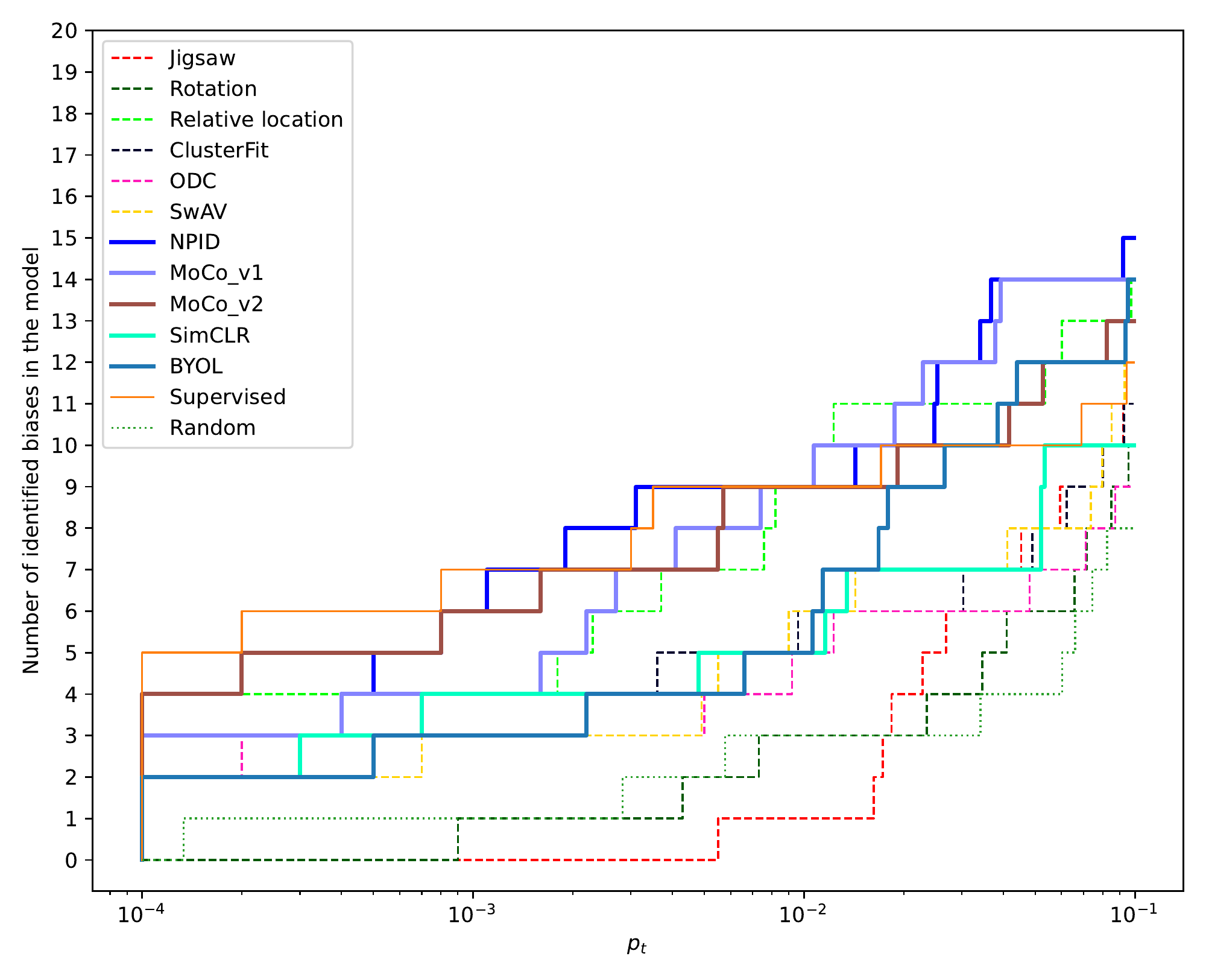}
       \caption{5\textsuperscript{th} ResNet block}
       \label{fig_2b}
  \end{subfigure}
  \caption{Number of biases for different values of $p_t$. Biases detected for lower values of $p_t$ are statistically more significant. Contrastive models are plotted with thick solid lines, geometric models and clustering-based models with dashed lines.}
  \label{fig_cumulative_layers_2_5}
\end{figure*}

\subsection{Per-layer analysis}



\begin{figure*}[t]
  \centering
  \begin{subfigure}[t]{0.5\textheight}
       \includegraphics[width=\textwidth]{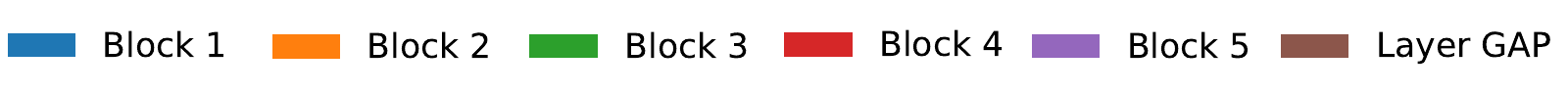}
  \end{subfigure}
    
  \begin{subfigure}[b]{0.47\textwidth}
       \includegraphics[width=\textwidth]{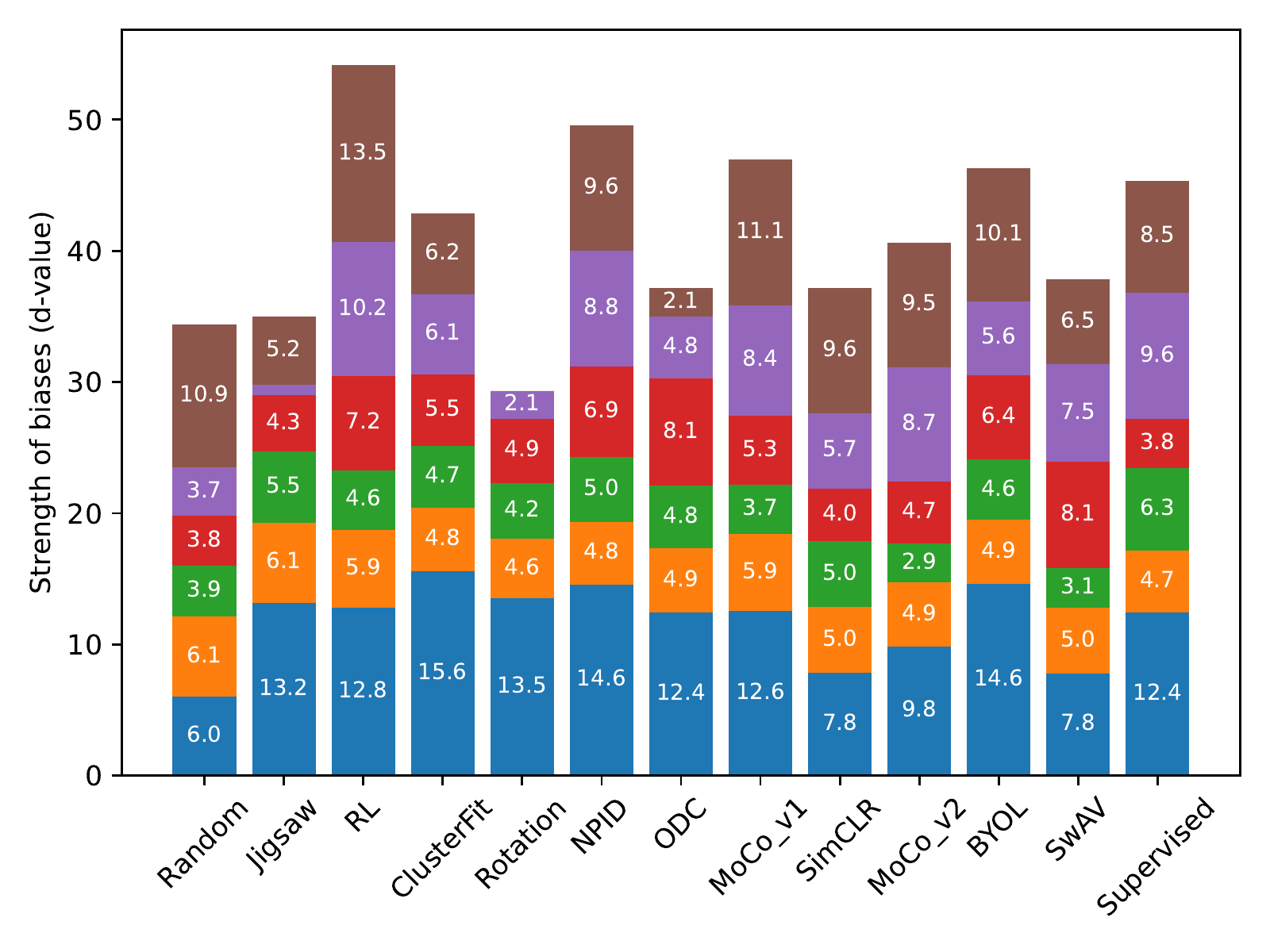}
       \caption{Cumulative strength of biases with $p$-value $<$ 0.01}
       \label{fig_3a}
  \end{subfigure}
  \hfill
  \begin{subfigure}[b]{0.47\textwidth}
       \includegraphics[width=\textwidth]{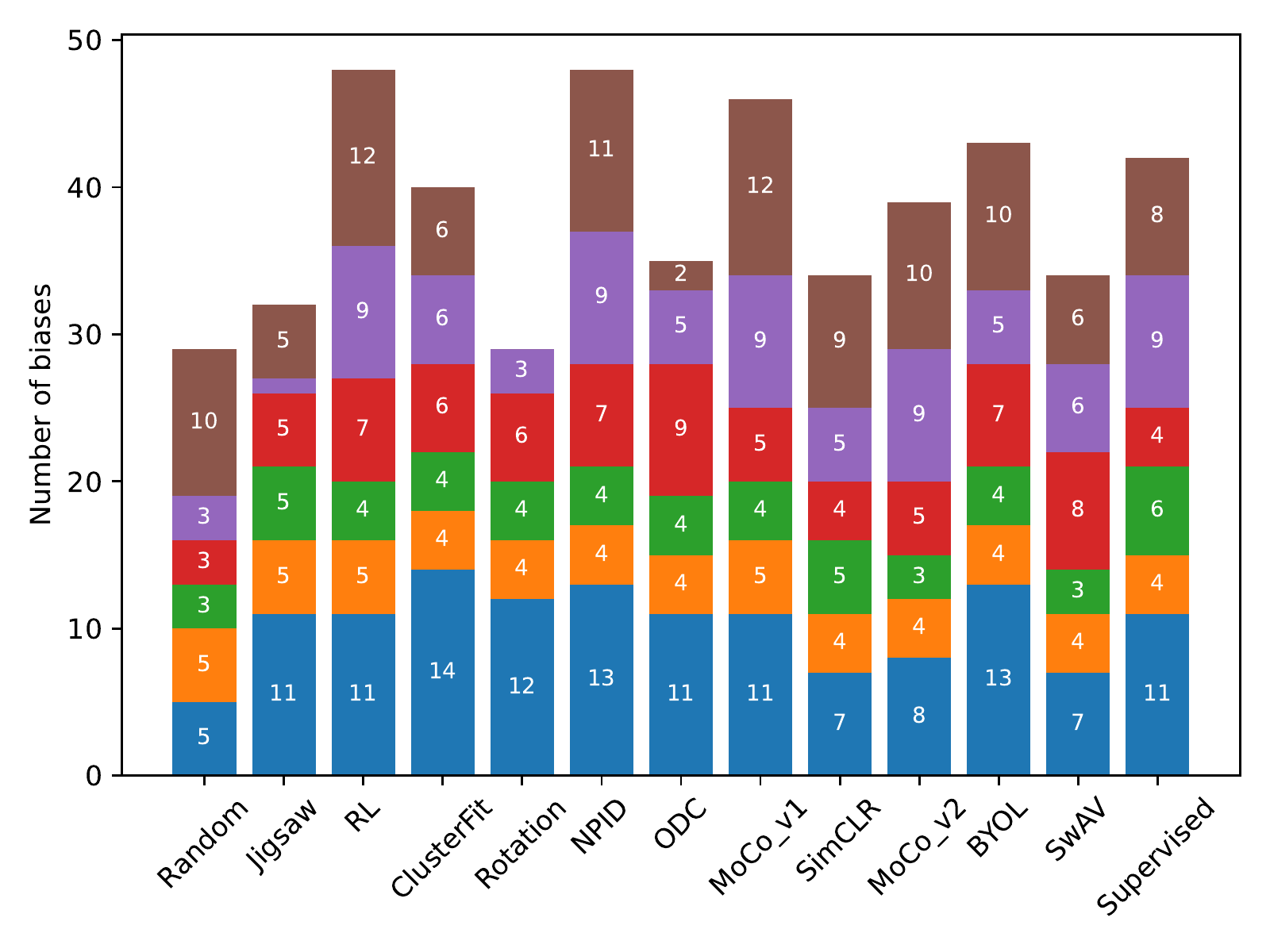}
       \caption{Number of biases with $p$-value $<$ 0.01}
       \label{fig_3b}
  \end{subfigure}
  \caption{Cumulative strength (on the left) and number (on the right) of biases detected in different layers. The number of biases correlates with the cumulative strength (see additional plots in supplementary materials). Models are ordered according to their classification accuracy on ImageNet.}
  \label{fig_cumulative_strength}
\end{figure*}

Anticipating that the strength and the number of biases varies for different layers of the network architecture due to increasing semantic interpretability in the internal CNN representations \cite{bib_network_dissection}, the bias-detection procedure is carried out on the feature embeddings extracted from all ResNet blocks. Our findings are partially depicted in Figure \ref{fig_cumulative_layers_2_5}, that presents the cumulative number of biases varying $p_t$ in the embeddings of the 2\textsuperscript{nd} and 5\textsuperscript{th} ResNet blocks. Figure \ref{fig_cumulative_strength} complements these results by summarizing the number and cumulative strength of biases in the embeddings extracted from all ResNet blocks. The strength of a bias refers to the $d$-value (Section \ref{section_methodology}), and the cumulative strength is the sum of the $d$-values of all detected biases.






The results shown on Figure~\ref{fig_2a} indicate that the biases are also detected in the feature embeddings of shallow ResNet layers that semantically resemble low-level features. However, the biases detected in the shallow layers mostly repeat for all models. For example, in Figure~\ref{fig_2a} four out of five biases at $p < 10^{-2}$ are common for all models and are race-related. On the other hand, the feature embeddings extracted from deeper layers, as shown on Figure~\ref{fig_2b}, result in more biases given the same value of the threshold. This statement holds for all the models except for the rotation prediction model. Model-wise, the number of biases with the same degree of statistical significance is more uniform for all models in the shallow layer embeddings and begins to differ towards the end of the network, with contrastive SSL and supervised models having a larger amount of biases.

Dissecting the bias detections at $p < 10^{-2}$ (Figure~\ref{fig_cumulative_strength}) for the embeddings of different ResNet blocks we gain insight into the distribution of the number and strength of the biases identified at different model depths. Overall, the cumulative strength of detected biases is smallest around the 3\textsuperscript{rd} and 4\textsuperscript{th} blocks, and it grows at the 5\textsuperscript{th} block. One can observe that some models (e.g., rotation, jigsaw and supervised) deviate from this pattern. Finally, for each model, the embeddings of the 1\textsuperscript{st} ResNet block yield some of the highest cumulative strength values. 

%% file: Discussion.tex
\section{Discussion on the experimental results} \label{section_discussion}

\subsection{Relation between the number of biases and SSL learning strategy}


Figures \ref{fig_cumulative_avg_pool} and \ref{fig_cumulative_strength} suggest that deep embeddings obtained with contrastive SSL models show more biases than the ones computed with geometric and clustering-based models. We hypothesize that the reason for this difference might lie in the nature of the contrastive loss function. A contrastive loss promotes the similarity between features of two images representing a concept and an attribute (not necessarily related) if the images are similar. Instead, a geometry-based loss function will not amplify this circumstantial similarity as much as a contrastive one.



Moreover, we state that a higher classification accuracy of the SSL model does not necessarily result in a higher number of social biases incorporated into the model. Figure \ref{fig_cumulative_strength} provides a good example of this by showing that one of the least accurate, among studied, model (relative location) yields the highest cumulative strength of biases detected in it. In addition to it, ODC demonstrates an inferior cumulative bias-strength than NPID, whilst being more accurate.


The aforementioned conclusions might have an important application in the deployment of deep learning models for tasks that have an impact on human processes: in the context of transfer learning, one must not solely rely on final accuracy when choosing a model or the layer-depth to extract embeddings from. Based on the ImageNet classification accuracies, we argue that it might be beneficial, for some models, to give preference to embeddings resulting in a slightly lower accuracy but significantly reducing the strength of identified biases. For instance, for two  linear classifiers trained on the embeddings from the 5\textsuperscript{th} block and the GAP layer of NPID, the difference in classification accuracy on ImageNet is only 0.01\%. Meanwhile, cumulative strength of biases of these two layers differs by 9\%. Furthermore, the classifier trained on the GAP embeddings of NPID is 3.18\% more accurate than the classifier trained on the GAP embeddings of ODC, but the number of \textit{intersectional} biases acquired by them differs significantly (see Figure \ref{fig_intersectional_bars}).


\subsection{Distribution of biases along the layers}

Although the presence of biases in initial layers of the network architecture might be counterintuitive, we believe that it can be explained through a correlation between the low-level characteristics of the test data and the nature of the filters learned in the shallow layers of CNNs, i.e., similar data issues affect the first layer and the random models as described in Section \ref{bias random}. For example, many of the biases that are consistently detected in the 1\textsuperscript{st} block embeddings of all models relate to the \textit{skin tone} and \textit{valence} or \textit{weight} and \textit{valence}. Considering that images representing the concept of \textit{pleasantness} contain brighter pixels (like the images of \textit{white skin tone}) and images representing the concept of \textit{unpleasantness} contain darker pixels (like the images of \textit{dark skin tone}), it could be expected that the correlation identified between corresponding embeddings might be caused by these data factors (as we explore in the supplementary materials) and not by the meaning of the depicted concepts. 

Regarding the distribution of the biases, besides the aforementioned behaviour in the first layer, biases grow in strength and quantity as one advances along the contrastive models, strongly correlating the number and intensity of the acquired biases with the classification potential of the embeddings at each layer of a given model\textemdash generally the deeper the better as reported in \href{https://github.com/open-mmlab/OpenSelfSup/blob/master/docs/MODEL_ZOO.md}{OpenSelfSupervised Framework}: the more specialized the embeddings are within a model, the more and stronger biases are acquired. The same trend is observable in the relative location model, and in a more subtle way in the ODC model. Biases in the supervised model are more evenly distributed along the layers and strongly increase in the last layers, maybe because it is the closest to the label-guided classification one. The rotation model shows a different behaviour with lower and less intense biases evenly distributed along the model without representative biases in the GAP layer. 

This analysis is complemented with an analysis on the nature of the acquired biases provided in the supplementary material.

%% file: Conclusion.tex
\section{Conclusion} \label{section_conclusions}

In this work, building on the existing approaches, we study the presence of common social biases in three types of SSL models: geometric, clustering-based and contrastive. We show that the number of detected biases does not depend on the SSL model's classification accuracy but on its type, with contrastive models yielding the highest number of biases. Moreover, we show that the presence of biases is not constant across different layers of a model, and that this layer-distribution of biases changes across models. Given these findings, we suggest that the number and strength of biases should be taken into account, alongside the resulting accuracy, when performing transfer learning on (supervised or SSL) pre-trained models. Specially for tasks that have an impact on human processes, this educated selection would result in models with a better trade-off in terms of accuracy and bias. Nevertheless, not all open questions have been answered yet: the sources of biases that stem from training data need to be isolated, and the influence of the dataset used during the models training needs to be investigated more closely. In fact, this study considered an ample number of models, although all of them trained only on ImageNet. This limitation opens avenues for further exploration of biases that arise in models trained using different datasets.

